\documentclass[11pt]{article}

\usepackage[utf8]{inputenc}
\usepackage[T1]{fontenc}
\usepackage{lmodern}

\usepackage{amsmath, amssymb}
\usepackage{graphicx}
\usepackage{booktabs}
\usepackage{float}
\usepackage{authblk}
\usepackage{tabularx}
\usepackage{array}
\usepackage{makecell}

\usepackage[hidelinks]{hyperref}

\newcolumntype{L}[1]{>{\raggedright\arraybackslash}p{#1}}
\newcolumntype{C}[1]{>{\centering\arraybackslash}p{#1}}

\newcommand{\keywords}[1]{\par\noindent\textbf{Keywords: }#1\par}

\title{Leveraging LLMs for Collaborative Ontology Engineering in Parkinson Disease Monitoring and Alerting}

\author[1]{Georgios Bouchouras\thanks{Corresponding author: \texttt{gbouchouras@aegean.gr}}}
\author[1]{Dimitrios Doumanas}
\author[1]{Andreas Soularidis}
\author[1]{Konstantinos Kotis\thanks{Corresponding author: \texttt{kotis@aegean.gr}}}
\author[2]{George A. Vouros}

\affil[1]{Intelligent Systems Laboratory, Department of Cultural Technology and Communication,\\
University of the Aegean, Mytilene 81100, Greece}

\affil[2]{Artificial Intelligence Laboratory, Department of Digital Systems,\\
University of Piraeus, Piraeus 18534, Greece}

\date{}

\begin{document}
\maketitle

\begin{abstract}
This paper explores the integration of Large Language Models (LLMs) in the engineering of a Parkinson's Disease (PD) monitoring and alerting ontology through four key methodologies: One Shot (OS) prompt techniques, Chain of Thought (CoT) prompts, X-HCOME, and SimX-HCOME+. The primary objective is to determine whether LLMs alone can create comprehensive ontologies and, if not, whether human-LLM collaboration can achieve this goal. Consequently, the paper assesses the effectiveness of LLMs in automated ontology development and the enhancement achieved through human-LLM collaboration.

Initial ontology generation was performed using One Shot (OS) and Chain of Thought (CoT) prompts, demonstrating the capability of LLMs to autonomously construct ontologies for PD monitoring and alerting. However, these outputs were not comprehensive and required substantial human refinement to enhance their completeness and accuracy.

X-HCOME, a hybrid ontology engineering approach that combines human expertise with LLM capabilities, showed significant improvements in ontology comprehensiveness. This methodology resulted in ontologies that are very similar to those constructed by experts.

Further experimentation with SimX-HCOME+, another hybrid methodology emphasizing continuous human supervision and iterative refinement, highlighted the importance of ongoing human involvement. This approach led to the creation of more comprehensive and accurate ontologies.

Overall, the paper underscores the potential of human-LLM collaboration in advancing ontology engineering, particularly in complex domains like PD. The results suggest promising directions for future research, including the development of specialized GPT models for ontology construction.
\end{abstract}


% ----------- The article body starts: -----------
\section{Introduction}
The integration of LLMs (Large Language Models) with ontological frameworks is gaining prominence in the fields of knowledge representation (KR) and Artificial Intelligence (AI) \cite{Chowdhery2022}. A noticeable trend is the use of LLMs for the construction, refinement, and mapping of ontologies, tasks traditionally performed and supervised by human experts with in-depth domain and ontology engineering knowledge, as KR methods become more demanding \cite{Uschold1996}. Training LLMs on big data makes expert-level insights across domains more accessible and cost-effective. Moreover, while LLMs are getting more effective at engineering ontologies \cite{Funk2023}, their capabilities are significantly enhanced in the era of Neurosymbolic AI, i.e., combining the deep and varied knowledge of statistical AI with the semantic reasoning of symbolic AI \cite{Sheth2023}.

Artificial intelligence is particularly significant in addressing complex health problems such as monitoring and alerting patients and doctors to Parkinson Disease (PD), the second most common neurodegenerative disease globally \cite{Corra2023}. Despite extensive research, the nature of PD remains elusive, and current treatments offer only partial effectiveness \cite{Bonuccelli2008}. In response, related ontologies have been developed to enhance understanding, monitoring, alerting, and treatment approaches. Specifically, the Wear4PDmove ontology \cite{Zafeiropoulos2023Wear4PDmove, Bitilis2023} has been recently developed with the aim of integrating heterogeneous sensor (movement) and personal health record (PHR) data, as a knowledge model used to interface/connect patients and doctors with smart devices and health applications. This ontology aims to semantically integrate heterogeneous data sources, such as dynamic/stream data from wearables and static/historic data from personal health records, to represent personal health knowledge in the form of a Personal Health Knowledge Graph (PHKG). Also, it supports health applications' reasoning capabilities for high-level event recognition in PD monitoring, such as identifying events like `missing dose' or `patient fall' \cite{Bitilis2023, Zafeiropoulos2023Wear4PDmove, Zafeiropoulos2023GNN}. This and associated ontologies facilitate the critical integration of domain-specific knowledge, making it easier to integrate and reason with health data and promoting PD treatment approaches.

Patients' PD monitoring and alerting requires flexible KR methods to effectively adapt to health changes. LLMs have demonstrated impressive abilities in handling large amounts of data and producing valuable insights from their near-real-time analysis. However, factors such as inadequate reasoning abilities and reliance on specialized health knowledge limit their use in monitoring PD and alerting patients. PD is a complex domain, with distinct contexts, subtle meaning variations, and disease-specific vocabularies. Effectively capturing and expressing this complex knowledge requires fine-tuning and training LLMs for the domain, demanding significant resources that are often unavailable or beyond the capacity of health and medical experts. Additionally, healthcare ontologies now adhere to several standards and forms. The technical challenge, however, lies in the integration and reconciliation of information from many heterogeneous sources into a coherent ontology, while also ensuring interoperability. To achieve an efficient ontology development process within an ontology engineering methodology (OEM), LLMs must be able to navigate these disparities efficiently. Existing research on PD exploits ontologies \cite{Zafeiropoulos2023GNN, Younesi2015}. However, maintaining these ontologies in this rapidly changing field of PD calls for constant effort and resources. Failure to update or refine the ontology may result in outdated information. This involves developing methods that streamline the ontology engineering process, making it more accessible and less resource-intensive.

Existing research has primarily focused on cooperation among participants, particularly domain experts collaborating with one another. However, real-time collaboration between humans and machines at various levels of participation in the development and improvement of ontologies using the OEM remains relatively underexplored. Notably, research has overlooked the extent of human involvement and the potential contribution of LLM assistance. Currently, many ontology engineers devote excessive time and resources to creating an initial ontology, known as the `kick-off ontology,' but they often lack effective automated methods for further development and refinement. It is crucial to examine the varying levels of contributions from both humans and machines throughout an OEM to demonstrate the methodology's comprehensiveness and the diversity of results, while aiming to save time and resources.

This paper defines varying levels of human involvement in LLM-based enhanced ontology engineering, corresponding to different OEMs. These levels range from minimal to moderate human involvement, allowing machines and humans to collaborate effectively. This transition moves from a human-centered to a more machine-centered OEM, with humans gradually transferring decision-making power to machines. This indicates an opportunity for developing new techniques to enhance ontologies, making them more time efficient and comprehensive.

In this paper, the authors introduce experiments for ontology engineering, towards engineering an ontology in the challenging PD domain. They also focus on expanding the human-centered collaborative OEM (HCOME) \cite{Kotis2006} through LLM-based tasks, a concept proposed and evaluated as X-HCOME. The authors additionally utilize another extension, the simulated X-HCOME (SimX-HCOME+), to further enhance and evaluate the methodology. This extension features simulated environments to test the interaction between human and machine contributions under controlled conditions, providing deeper insights into the dynamics of collaborative OEM. The aim is to provide a novel OEM, including both humans and LLMs in the engineering of ontologies, with a focus on comprehensiveness and conciseness of conceptualizations, and the required level. The final product of this is an OEM constructing domain ontologies more effectively and with time efficiency than those used solely by humans or LLMs. The paper focuses on LLM-based collaborative OEM to create comprehensive PD ontologies and discusses findings and limitations of the LLM-based collaborative process, identified from the experimental results.

Building upon previously published research \cite{BouchourasGeorgiosBitilisPavlosKotisKonstantinos2024} this paper studies several significant extensions: (a) the implementation and evaluation of a new methodology for LLM-enhanced ontology engineering (SimX-HCOME+); (b) the addition of a new capability of the proposed approach to convert a rule from natural language (NL) to Semantic Web Rule Language (SWRL); and (c) a comparison of the highest LLM performance and the degree of human involvement, across all methodologies. These contributions aim to improve the comprehensiveness of the human-LLM generated ontology for PD monitoring and alerting.

This paper's organization is as follows: Section 2 presents related work on integrating LLMs into OEM; Section 3 describes the proposed research methodology and hypotheses; Section 4 presents the results of the conducted experiments; Section 5 presents a comprehensive evaluation comparing the performance of LLMs across various methodologies, highlighting the degree of human involvement in each approach; finally, Section 6 discusses the results and draws conclusions.

\section{Related Work}
Oksannen et al. (2021) developed an approach to derive product ontologies from textual reviews using BERT models. Their approach, which required minimum manual annotation, demonstrates increased precision and recall in comparison to established methods such as Text2Onto and COMET, signifying a noteworthy advancement in automatic ontology extraction \cite{Oksanen2021}. The BERTMap, a tool designed for the visualization and analysis for Bidirectional Encoder Representations from Transformers by He et al. (2022), demonstrates the effectiveness of LLMs by excelling at ontology mapping (OM), especially in unsupervised and semi-supervised scenarios, surpassing current OM systems. It demonstrates the precision of LLMs in matching entities between knowledge graphs \cite{He2022}. Ning et al. (2022) introduce a technique to extract factual information from LLMs by creating prompts for pairs of subjects and relations. They utilize an approach that incorporated pre-trained LLMs with prompt templates derived from web material and personal expertise. The authors identify effective prompts through a parameter selection technique and filter the generated entities to pinpoint reliable choices. They stress the significance of investigating parameter combinations, testing LLMs, and expanding research into different domains \cite{Ning2022}.

Lippolis et al. (2023) concentrate on harmonizing entities across ArtGraph and Wikidata. By combining traditional querying with LLMs, they achieve a high accuracy in entity alignment, showcasing the efficiency of LLMs in filling knowledge gaps in intricate databases \cite{Lippolis2023}. Funk et al. (2023) investigates the capability of GPT3.5 (Generative Pre-trained Transformer) in creating concept hierarchies in several fields. Their method decreases mistakes and generates appropriate concept names, demonstrating the effectiveness of LLMs in the semi-automatic creation of ontologies. Studies on GPT4's abilities in structured intelligence within ontologies indicate its potential for groundbreaking progress. Their study emphasizes the importance of implementing controlled LLM integration in business environments through a collaborative framework \cite{Funk2023}. Biester et al. (2023) develops a technique that utilizes prompt ensembles to improve knowledge base development. When applied to models such as ChatGPT and Google BARD, they demonstrate notable enhancements in precision, recall, and F-1 score, highlighting the effectiveness of LLMs in improving knowledge bases \cite{Biester2023}. Mountantonakis and Tzitzikas (2023) devise a technique to verify ChatGPT information by utilizing RDF Knowledge Graphs. They confirm the accuracy of 85.3\% of ChatGPT facts, highlighting the significance of verification services in maintaining data precision \cite{Mountantonakis2023}. Pan et al. (2023) suggests combining LLMs with KGs to improve reasoning skills. Their frameworks attempt to combine the benefits of both LLMs and KGs, resulting in enhanced data processing and reasoning abilities \cite{Pan2023}. Joachimiac et al. (2023) used the Spindoctor approach, which employed LLMs to summarize gene sets, demonstrating the versatility of LLMs in analyzing intricate biological information. Their method showcased the effectiveness of LLMs in summarizing text specifically related to gene ontology \cite{Joachimiak2009}. The SPIRES approach developed by Caufield et al. (2023) demonstrates the adaptability of LLMs in extracting information from unstructured texts in many fields. This zero-shot learning method does not require any model adjustment, demonstrating the wide range of applications of LLMs in various disciplines \cite{Caufield2023}. Mateiu et al. (2023) showcase the application of GPT3 in converting natural language words into ontology axioms. Their methodology facilitates ontology creation, enhancing accessibility and efficiency, demonstrating the effectiveness of LLMs in streamlining intricate ontology engineering processes \cite{Mateiu2023}.

However, the aforementioned studies primarily concentrate on the capabilities of LLMs in isolation or in comparison with traditional methods, often emphasizing automated or semi-automated processes. What remains less explored, and thus the focus of current paper, is the integration of human expertise and LLMs capabilities in the process of OEM. This novel approach aims to harness the large corpus of knowledge, speed in shaping results of LLMs while simultaneously capitalizing on the complex understanding and conceptualization skills of human experts. Furthermore, it is reasonable to believe that the differences between LLMs have strengths and weaknesses that can help researchers and practitioners choose the best models for use in real-world entity resolution \cite{Zeakis2023}.

\section{Research Methodology}
This section presents experiments encompassing four distinct phases, focusing on the development and assessment of ontologies, with a special emphasis on classes. The initial phase involves generating an ontology for PD monitoring and alerting, mainly powered by the capabilities of LLMs. This process utilizes both `One Shot' (OS) and `Chain of Thought' (CoT) techniques. The OS method involves presenting a model with a single prompt and expecting it to produce a suitable response based only on this input. In a one-shot scenario, the model lacks multiple learning examples and must accomplish the task with minimal context. This is a straightforward approach where the model uses its pre-trained knowledge to infer the most likely answer. For the purposes of this paper, CoT refers to a methodological approach where the OS is segmented into two sequential prompts. This segmentation allows for a structured progression in the reasoning process, whereby each prompt is designed to focus on a specific element of the overall task. By employing sequential prompting, the authors direct the language model to tackle each segment of the problem individually, thereby facilitating a cumulative build-up of information.

\begin{figure}[H]
  \centering
  \includegraphics[width=0.75\linewidth]{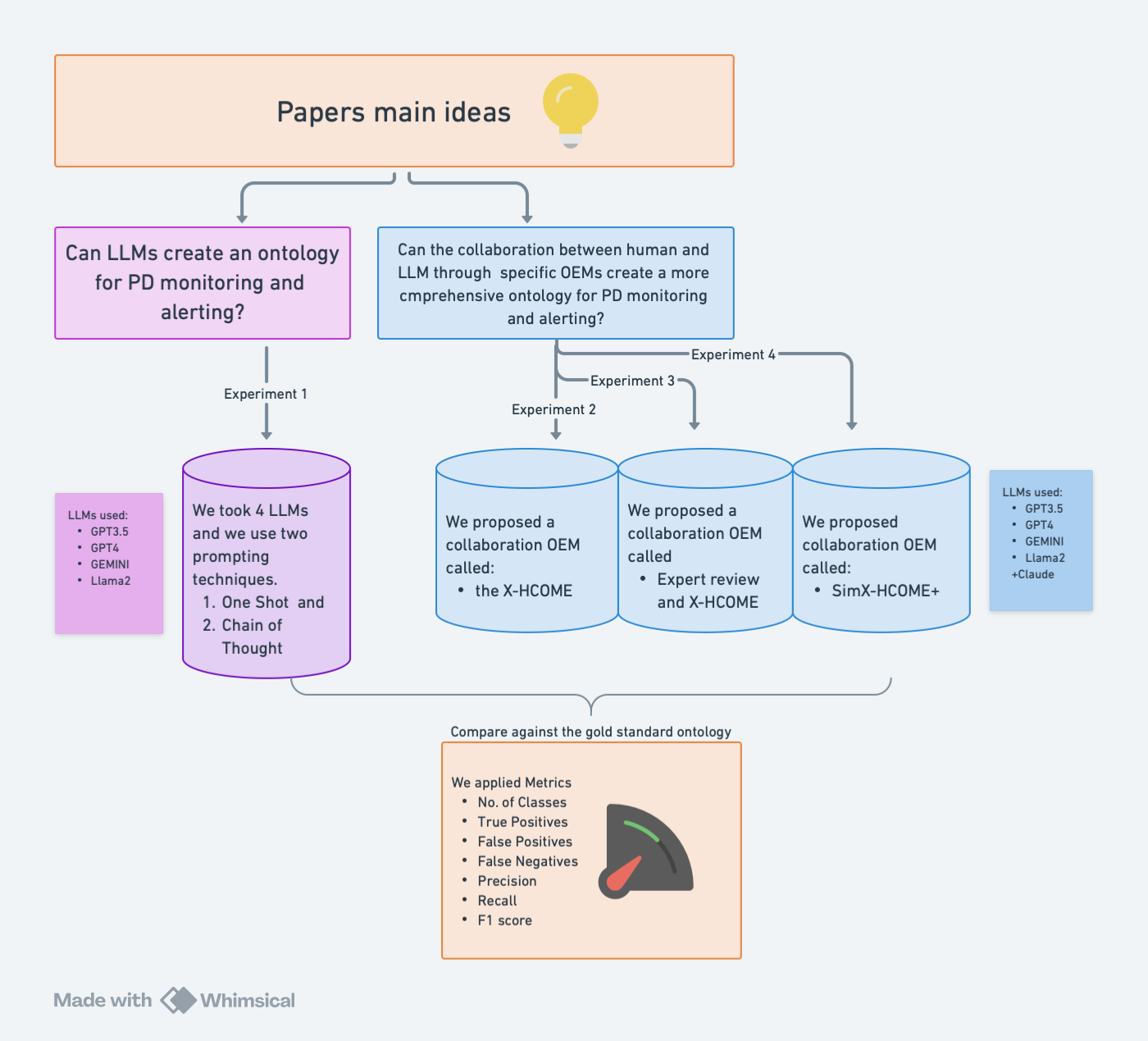}
  \caption{Flowchart of a multi-phase experimentation assessing the construction and validation of ontologies using different methodologies (created with AI-Whimsical ChatGPT, 2023).}
  \label{fig:exp-flow}
\end{figure}

Subsequently, in following experiments, hybrid OEMs are established, which integrate human expertise with the abilities of LLMs. This collaboration aims to elevate the comprehensiveness of the ontology within the PD monitoring and alerting framework. Figure~\ref{fig:exp-flow} depicts a flowchart that outlines this four-phase experimental process. Initially, four LLMs independently develop an ontology with minimal human input (Experiment 1). The process evolves into a more collaborative approach (human and LLMs; Experiments 2--4). The authors compare the resulting ontologies against a gold standard ontology. In this paper, the Wear4PDmove ontology \cite{Zafeiropoulos2023Wear4PDmove, Bitilis2023} is utilized as the gold standard ontology, and it will be referred to as such throughout the remainder of the paper.

\footnotetext{OpenAI. 2023. ``Whimsical Diagrams.'' ChatGPT functionality. \url{https://openai.com/chatgpt}.}

\textbf{Hypothesis 1:} LLMs, when prompted with domain-specific queries, can autonomously develop a comprehensive ontology, as it is in the case of PD monitoring and alerting ontology. LLMs have the ability to extract domain knowledge efficiently from their extensive corpus of domain knowledge, and construct ontologies using different prompts provided by humans. This hypothesis is tested in Experiment 1, where LLMs are tasked with creating a PD monitoring and alerting ontology from ground zero, using domain-specific prompts. The effectiveness of LLMs in developing an accurate and relevant ontology is measured against the gold standard ontology.

\footnotetext{\url{https://oops.linkeddata.es}.}
\footnotetext{\url{https://protege.stanford.edu}.}

\textit{Experiment 1: Initiating LLMs to develop the ontology.} During the initial phase of the experiments, the LLMs will independently (minimum human-involvement) construct an ontology for PD monitoring and alerting from scratch. This phase comprises the following steps:
\begin{enumerate}
    \item LLMs construct an ontology in Turtle format. The ontology represents various aspects of PD patient care, including monitoring, alerting, patients' health record and healthcare team coordination.
    \item Validate the ontology by assessing its consistency with OOPS! and Prot\'eg\'e tools (Pellet Reasoner).
    \item Use metrics such as Precision, Recall, and the F-1 score (Table~\ref{tab:metrics}) to compare the LLM-generated ontology comprehensiveness against the gold standard ontology.
\end{enumerate}

\begin{table}[H]
\centering
\caption{Summary of metrics for class evaluation. This table presents the formulas for Precision, Recall, and the F1-score, along with their definitions.}
\label{tab:metrics}
\small
\setlength{\tabcolsep}{4pt}
\renewcommand{\arraystretch}{1.15}

\begin{tabularx}{\textwidth}{|p{0.40\textwidth}|X|}
\hline
\textbf{Formulas} & \textbf{Definitions} \\
\hline
$\displaystyle \text{Precision}=\frac{\text{TP}}{\text{TP}+\text{FP}}$ &
True Positives (TP): classes correctly classified as positive in alignment with the gold standard ontology (human judgment or alignment tool). \\
\hline
$\displaystyle \text{Recall}=\frac{\text{TP}}{\text{TP}+\text{FN}}$ &
False Positives (FP): classes incorrectly classified as positive in alignment with the gold standard ontology. \\
\hline
$\displaystyle \text{F1}=2\cdot\frac{\text{Precision}\cdot\text{Recall}}{\text{Precision}+\text{Recall}}$ &
False Negatives (FN): classes incorrectly classified as negative despite being positive in the gold standard ontology. \\
\hline
\end{tabularx}
\end{table}

The methods in this section and the results listed below, supported by supplementary material placed at a GitHub repository,\footnote{\url{https://github.com/GiorgosBouh/Ontologies_by_LLMst}.} focus on the complex process of creating ontologies for monitoring and alerting patients in PD. It is essential to clarify that the metrics presented in this paper are solely focused on the generated ontological classes and object properties. The validation involves both exact matching, where generated entities corresponded to entities in the gold standard ontology, and similarity matching, where entities are considered correct if they were semantically similar to the gold standard ones. Exact matching quantifies direct accuracies, while similarity matching captures the broader context and appropriateness of the generated entities. This approach aims to provide a comprehensive evaluation of the LLM's performance, capturing both direct accuracies and contextually appropriate approximations.

LLMs are initially given prompts with two methods. The one-shot prompting (OS) method provided the LLMs with a single, clear prompt that clearly stated the aim and scope of the gold standard ontology, without any additional information or background. The goal was to test LLMs' initial response effectiveness by generating accurate and relevant ontologies from a single standalone prompt, with minimal human effort.

The following paragraph provides an example of an OS prompt:
\textit{``Act as an Ontology Engineer, I need to generate an ontology about Parkinson disease monitoring and alerting patients. The aim of the ontology is to collect movement data of Parkinson disease patients through wearable sensors, analyze them in a way that enables the understanding (uncover) of their semantics, and use these semantics to semantically annotate the data for interoperability and interlinkage with other related data. You will reuse other related ontologies about neurodegenerative diseases. In the process, you should focus on modeling different aspects of PD, such as disease severity, movement patterns of activities of daily living, and gait. Give the output in TTL format.''}

\textbf{Chain-of-Thought prompting (CoT):} The CoT prompting method breaks down the OS prompt into two distinct prompts as follows:

Prompt 1: \textit{``Act as an Ontology Engineer, I need to generate an ontology about Parkinson disease monitoring and alerting patients. The aim of the ontology is to collect movement data of Parkinson disease patients through wearable sensors, analyze them in a way that enables the understanding (uncover) of their semantics, and use these semantics to semantically annotate the data for interoperability and interlinkage with other related data.''}

Prompt 2: \textit{``You will reuse other related ontologies about neurodegenerative diseases. In the process, you should focus on modeling different aspects of PD, such as disease severity, movement patterns of activities of daily living and gait. Give the output in TTL format.''}

The first prompt covers the role, aim, and scope of the ontology and sets the foundation for the ontology. The second prompt deals with the processing and utilization of the data collected as per the framework set up in the first prompt.

\textbf{Hypothesis 2:} The combination of human expertise and LLM capabilities enhances the comprehensiveness of the developed ontology, as it is in the case of PD monitoring and alerting ontology.

This hypothesis is related to the second experimentation and specifies a new OEM called `X-HCOME'. The X-HCOME methodology is an extension of the Human-Centered Collaborative Ontology Engineering methodology (HCOME) \cite{Kotis2006}. This extension concerns the inclusion of LLM-based tasks (along with human-centered ones) in the OE lifecycle. It assesses how the collaboration between humans and LLMs contributes to refining and validating the ontology, ensuring its relevance and accuracy, e.g., in the case of PD monitoring and alerting patients. The X-HCOME methodology is a novel approach in OE that integrates the expertise of human experts (domain expert and ontology engineer) with the computational power of LLMs in domain knowledge acquisition and ontology engineering. At each stage of this iterative process, human domain experts critically examine and provide feedback on the ontologies generated by the LLMs.

\textit{Experiment 2.} The X-HCOME methodology that this paper presents involves a number of steps assigned to either human experts or LLMs in an alternating manner during the OE process:
\begin{enumerate}
    \item (Human): Define prompts and provide LLMs with the specified data. (a) Define the aim and scope of the ontology. (b) Ontology requirements. (c) Integrate data from PD cases. (d) Formulate competency questions in natural language.
    \item (LLM): Construct a domain ontology using the input provided by the human (e.g., Turtle).
    \item (Human): Compare the LLM-generated ontology with existing gold standard ontologies, manually or assisted by OM tools (e.g., LogMap \cite{Jimenez-Ruiz2011}).
    \item (LLM): Perform a machine-based comparison of LLM-generated ontology against the gold standard ontology (LLM-assisted OM).
    \item (Human): Develop a revised domain ontology by combining an existing ontology with the one generated by the LLM.
    \item (LLM): Repeat machine-based evaluation of the developed ontology.
    \item (Human): Evaluate the revised/refined ontology using OE tools (consistency/validity checks).
\end{enumerate}

\textbf{Hypothesis 3:} Analyzing false positives and incorporating domain expert opinions, LLMs can identify relevant domain knowledge not included in the gold standard ontology.

\textit{Experiment 3.} The expert review of the X-HCOME.
To enhance the evaluation of the generated ontologies, the authors conducted an in-depth analysis of the false positives. Acting as domain experts, they assessed whether the LLMs provided domain knowledge that the gold standard ontology might have missed due to human bias or other limitations in the original engineering. The goal of this analysis was to determine if the LLM-generated entities could be reclassified as true positives, even though they did not match entities in the gold standard ontology.

\textbf{Hypothesis 4:} Simulated collaboration between human experts and LLMs enhances ontology engineering by introducing a methodology where LLMs lead ontology development tasks within a controlled environment.

\textit{Experiment 4.} This newly introduced methodology, named Simulated X-HCOME (SimX-HCOME+), allows LLMs to leverage their strengths in NLP and knowledge extraction to autonomously build ontologies under human supervision. SimX-HCOME+ introduces a simulated environment where LLMs take the lead in ontology development tasks, but under the supervision of human experts. An iterative conversation between the three main roles Knowledge Worker (KW), Domain Expert (DE), and Knowledge Engineer (KE) is simulated. The approach incorporates continuous ontology generation and refinement throughout the iterative process. It ensures that ontologies are produced at every step, allowing for more comprehensive and detailed results.

During the OE lifecycle, the LLM user (human) feeds the LLM with related data, such as the aim and scope of the ontology, competency questions, and prompts the model to perform the LLM tasks defined in X-HCOME. When the collaborative and iterative execution of OE tasks ends, the final outcome (generated ontology) is delivered and evaluated.

The authors assessed the accuracy of LLM in properly identifying the number of entities, as well as its capability to transform rules from natural language (NL) to Semantic Web Rule Language (SWRL). The requested rule for the LLMs to generate and locate in the gold ontology was as follows: ``If an observation indicates that there is bradykinesia of the upper limb (indicating slow movement) and this observation pertains to the property and the observation is made after medication dosing, then a notification should be sent indicating a \texttt{<MissingDoseNotification>} and this observation should be marked as a \texttt{<PDpatientMissingDoseEventObservation>}.''

\section{Results}
\subsection{Experiments 1 and 2 (OS, CoT and X-HCOME)}
Ontological class definition consistency and syntactical correctness were observed in all LLM and collaborative generated ontologies, apart from the ones generated by Llama2 (OS, CoT and X-HCOME). The ontologies generated by Llama2 contained both syntactical errors and inconsistent definitions, which hindered its ability to produce a valid ontology. Also, all the developed ontologies were validated with OOPS!, identifying only one minor pitfall (pitfall P36-URI, file extension) during the experimental process.

Based on the data provided in Table 2, the chatGPT3.5 OS method identified 5 classes but had relatively low accuracy (precision 40\%, recall 5\%, F-1 score 9\%). ChatGPT3.5 CoT achieved higher precision (67\%) with limited recall (5\%), identifying only 3 classes. ChatGPT4 OS improved, identifying 9 classes (precision 56\%, recall 12\%, F-1 score 20\%), while ChatGPT4 CoT showed further enhancement with 6 classes (precision 67\%, recall 10\%, F-1 score 17\%). Conversely, Bard/Gemini OS had lower precision (8\%) and recall (2\%), identifying 13 classes, whereas Bard/Gemini CoT identified 8 classes with better precision (63\%) and recall (12\%), mirroring ChatGPT4 OS's performance.

For the X-HCOME method, the ChatGPT3.5 X-HCOME generated 25 classes with a precision of 40\%, a recall of 24\%, and an F-1 score of 30\%. The ChatGPT4 X-HCOME generated 33 classes but with lower precision (30\%), a recall of 24\%, and an F-1 score of 27\%. Remarkably, the Bard/Gemini X-HCOME method produced the highest number of classes (50) with a precision of 38\%, a recall of 46\%, and an F-1 score of 42\%. The Llama2 results indicated syntactical errors.

Overall, the X-HCOME methodology performed better in all LLMs. The Bard/Gemini X-HCOME method appeared to be the most effective overall in the context of ontology creation.

As for the object properties, the F-1 score across all methods varied from 6\% to 12\% indicating low performance. For the ChatGPT3.5 CoT, ChatGPT3.5 OS, and BARD/Gemini OS methods, the F1 score was 0\%. The complete table of results for object properties is in the GitHub repository.\footnote{\url{https://github.com/GiorgosBouh/Ontologies_by_LLMst}.}

\begin{table}[H]
\centering
\caption{Comparative evaluation of methodologies used for ontology creation against the gold standard ontology.}
\resizebox{\textwidth}{!}{
 \begin{tabular}{lccccccc}
\toprule
\textbf{Method} & \textbf{Number of Classes} & \textbf{True Positives} & \textbf{False Positives} & \textbf{False Negatives} & \textbf{Precision} & \textbf{Recall} & \textbf{F-1 score} \\
\midrule
Gold-ontology & 41 & & & & & & \\
\hline
ChatGPT3.5 CoT & 3 & 2 & 1 & 39 & 67\% & 5\% & 9\% \\
\hline
ChatGPT3.5 OS & 5 & 2 & 3 & 39 & 40\% & 5\% & 9\% \\
\hline
ChatGPT3.5 X-HCOME & 25 & 10 & 15 & 31 & 40\% & 24\% & 30\% \\
\hline
ChatGPT4 CoT & 6 & 4 & 2 & 37 & 67\% & 10\% & 17\% \\
\hline
ChatGPT4 OS & 9 & 5 & 4 & 36 & 56\% & 12\% & 20\% \\
\hline
ChatGPT4 X-HCOME & 33 & 10 & 23 & 31 & 30\% & 24\% & 27\% \\
\hline
Bard/Gemini CoT & 8 & 5 & 3 & 36 & 63\% & 12\% & 20\% \\
\hline
Bard/Gemini OS & 13 & 1 & 12 & 40 & 8\% & 2\% & 4\% \\
\hline
Bard/Gemini X-HCOME & 50 & 19 & 31 & 22 & 38\% & 46\% & 42\% \\
\hline
Llama2 CoT & 3 & 3 & 0 & 38 & 100\% & 7\% & 14\% \\
\hline
Llama2 OS & 2 & 2 & 0 & 39 & 100\% & 5\% & 9\% \\
\hline
Llama2 X-HCOME & 32 & 4 & 28 & 37 & 13\% & 10\% & 11\% \\
\bottomrule
\end{tabular}
}
\end{table}

\subsection{Experiment 3 (Expert review of the X-HCOME)}
The ChatGPT3.5 CoT and OS methods have comparable results, with the CoT method showing slightly higher precision but equal recall and an F-1 score as OS. For ChatGPT4, both CoT and OS showed similar trends, with CoT slightly outperforming OS in precision and recall (Table 3). Specifically, for ChatGPT3.5, X-HCOME significantly improved metrics for classes, achieving 92\% precision, 56\% recall, and a 70\% F-1 score, compared to lower scores for CoT and OS. ChatGPT4 showed similar trends, with X-HCOME achieving 88\% precision, 71\% recall, and a 78\% F-1 score. Bard/Gemini’s X-HCOME method excelled, showing no false positives and a high true positive rate, with an F-1 score of 110\%. Values above 100\% suggest that the true positives reported exceed the actual positives in the gold ontology.

For instance, Bard/Gemini X-HCOME generated classes like ``Surgical Intervention,'' ``Rigidity,'' and ``Cognitive Impairment,'' that were absent in the gold standard ontology. These classes enhance the ontology's utility in developing sophisticated PD monitoring and alerting systems.

As for the object properties, the F-1 score across all LLMs varied from 6\% to 84\%. All related metrics for object properties are presented at the GitHub repository.\footnote{\url{https://github.com/GiorgosBouh/Ontologies_by_LLMs}.}

\begin{table}[H]
\centering
\caption{Comparative evaluation of ontology creation methods’ post expert review on False Positives.}
\resizebox{\textwidth}{!}{
 \begin{tabular}{lccccccc}
\toprule
\textbf{Method} & \textbf{Number of Classes} & \textbf{True Positives} & \textbf{False Positives} & \textbf{False Negatives} & \textbf{Precision} & \textbf{Recall} & \textbf{F-1 score} \\
\midrule
Gold-ontology & 41 & & & & & & \\
\hline
ChatGPT3.5 CoT & 3 & 2 & 1 & 39 & 67\% & 5\% & 9\% \\
\hline
ChatGPT3.5 OS & 5 & 2 & 3 & 39 & 40\% & 5\% & 9\% \\
\hline
ChatGPT3.5 X-HCOME & 25 & 23 & 2 & 18 & 92\% & 56\% & 70\% \\
\hline
ChatGPT4 CoT & 6 & 4 & 2 & 37 & 67\% & 10\% & 17\% \\
\hline
ChatGPT4 OS & 9 & 5 & 4 & 36 & 56\% & 12\% & 20\% \\
\hline
ChatGPT4 X-HCOME & 33 & 29 & 4 & 12 & 88\% & 71\% & 78\% \\
\hline
Bard/Gemini CoT & 8 & 5 & 3 & 36 & 63\% & 12\% & 20\% \\
\hline
Bard/Gemini OS & 13 & 1 & 12 & 40 & 8\% & 2\% & 4\% \\
\hline
Bard/Gemini X-HCOME & 50 & 50 & 0 & -9 & 100\% & 122\% & 110\% \\
\hline
Llama2 CoT & 3 & 3 & 0 & 38 & 100\% & 7\% & 14\% \\
\hline
Llama2 OS & 2 & 2 & 0 & 39 & 100\% & 5\% & 9\% \\
\hline
Llama2 X-HCOME & 32 & 26 & 6 & 15 & 81\% & 63\% & 71\% \\
\bottomrule
\end{tabular}
}
\end{table}

\subsection{Experiment 4 (SimX-HCOME+)}
For SimX-HCOME, the evaluation criteria include ontology reusability, consistency (using Pellet Reasoner), syntactical errors, and whether the ontology can be opened by Prot\'eg\'e. ChatGPT4, ChatGPT3.5, and Claude all achieved ontology reusability, consistency without syntactical errors, and could be opened by Prot\'eg\'e. Gemini, while reusing ontology and being editable by Prot\'eg\'e, had syntactical errors.

The evaluation metrics of the SimX-HCOME+ generated ontologies in the PD domain reveal varying performances among the methods used (Table 4). Finally, the authors evaluated object properties, but due to space limitations, these results are not presented here. The full results for the object properties are in the GitHub repository.\footnote{\url{https://github.com/GiorgosBouh/Ontologies_by_LLMs}.}

\begin{table}[H]
\centering
\caption{Evaluation metrics on SimX-HCOME+ generated ontologies in PD domain (classes).}
\resizebox{\textwidth}{!}{
 \begin{tabular}{lccccccc}
\toprule
\textbf{Method} & \textbf{Number of Classes} & \textbf{True Positives} & \textbf{False Positives} & \textbf{False Negatives} & \textbf{Precision} & \textbf{Recall} & \textbf{F-1 Score} \\
\midrule
Gold ontology & 41 & & & & & & \\
\hline
ChatGPT-4 & 17 & 9 & 8 & 32 & 52\% & 21\% & 31\% \\
\hline
ChatGPT-3.5 & 21 & 14 & 7 & 27 & 66\% & 34\% & 45\% \\
\hline
Gemini & 22 & 15 & 7 & 26 & 68\% & 36\% & 48\% \\
\hline
Claude & 24 & 12 & 12 & 29 & 50\% & 29\% & 37\% \\
\bottomrule
\end{tabular}
}
\end{table}

Regarding the SWRL rules, while all LLMs except Gemini were able to generate the correct SWRL format, only a small number of logical atoms were detected, resulting in low performance and metrics. Among them, Claude had slightly better results (Table 5).

\begin{table}[H]
\centering
\caption{Evaluation metrics on SimX-HCOME+ generated ontologies in PD domain (NL2SWRL) with SC: syntactical comparison and LC: logical comparison.}
\resizebox{\textwidth}{!}{
\begin{tabular}{L{3cm} C{1.2cm} C{1.2cm} C{1.2cm} C{1.2cm} C{1.2cm} C{1.2cm} C{1.2cm} C{1.2cm} C{1.2cm} C{1.2cm} C{1.2cm} C{1.2cm} C{1.2cm}}
\toprule
\makecell{\rotatebox[origin=c]{90}{\textbf{Method}}} &
\makecell{\rotatebox[origin=c]{90}{\textbf{Atoms}}} &
\makecell{\rotatebox[origin=c]{90}{\textbf{TP SC}}} &
\makecell{\rotatebox[origin=c]{90}{\textbf{TP LC}}} &
\makecell{\rotatebox[origin=c]{90}{\textbf{FP SC}}} &
\makecell{\rotatebox[origin=c]{90}{\textbf{FP LC}}} &
\makecell{\rotatebox[origin=c]{90}{\textbf{FN SC}}} &
\makecell{\rotatebox[origin=c]{90}{\textbf{FN LC}}} &
\makecell{\rotatebox[origin=c]{90}{\textbf{Prec SC (\%)}}} &
\makecell{\rotatebox[origin=c]{90}{\textbf{Prec LC (\%)}}} &
\makecell{\rotatebox[origin=c]{90}{\textbf{Rec SC (\%)}}} &
\makecell{\rotatebox[origin=c]{90}{\textbf{Rec LC (\%)}}} &
\makecell{\rotatebox[origin=c]{90}{\textbf{F1 SC}}} &
\makecell{\rotatebox[origin=c]{90}{\textbf{F1 LC}}} \\
\midrule
Gold ontology & 8 & & & & & & & & & & & & \\
\hline
ChatGPT-4 & 13 & 0 & 3 & 13 & 10 & 8 & 5 & 0 & 23 & 0 & 27 & 0\% & 13\% \\
\hline
ChatGPT-3.5 & 17 & 1 & 3 & 16 & 14 & 7 & 5 & 5 & 17 & 12.5 & 3 & 1\% & 11\% \\
\hline
Gemini & 0 & 0 & 0 & 0 & 0 & 0 & 0 & 0 & 0 & 0 & 0 & 0\% & 0\% \\
\hline
Claude & 12 & 0 & 5 & 12 & 7 & 8 & 3 & 0 & 41.6 & 0 & 28.4 & 0\% & 20\% \\
\bottomrule
\end{tabular}
}
\end{table}

\section{Levels of Human Involvement Across Different Methodological Approaches in OE}
All the methodological approaches introduced in this paper align with distinct levels of human-machine collaboration, forming a spectrum from human-centered to LLM-centered collaborative ontology engineering. Within this spectrum, this section presents each methodology proposed. The authors arbitrarily allocated the degrees of human involvement to assess and compare the impact of varying levels of human participation on the ontology engineering process. The authors created a scale from 1 to 5 (with respect to LLM participation) to assess the different levels of human involvement (Table 6). This arbitrary assignment enables controlled analysis, guaranteeing consistent interpretation of the results across various methodological approaches.

\begin{table}[H]
\centering
\caption{Levels of Human Involvement Across Different Methodological Approaches in Ontology Engineering}
\resizebox{\textwidth}{!}{%
\begin{tabular}{lccccc}
\hline
\textbf{Methodological Approach} &
\textbf{OS} &
\textbf{CoT} &
\textbf{SimX-HCOME+} &
\textbf{X-HCOME} &
\textbf{Expert Review X-HCOME} \\
\hline
\textbf{Level of Human Involvement} & 1 & 2 & 3 & 4 & 5 \\
\hline
\end{tabular}
}
\end{table}

\begin{figure}[H]
    \centering
    \includegraphics[width=0.75\linewidth]{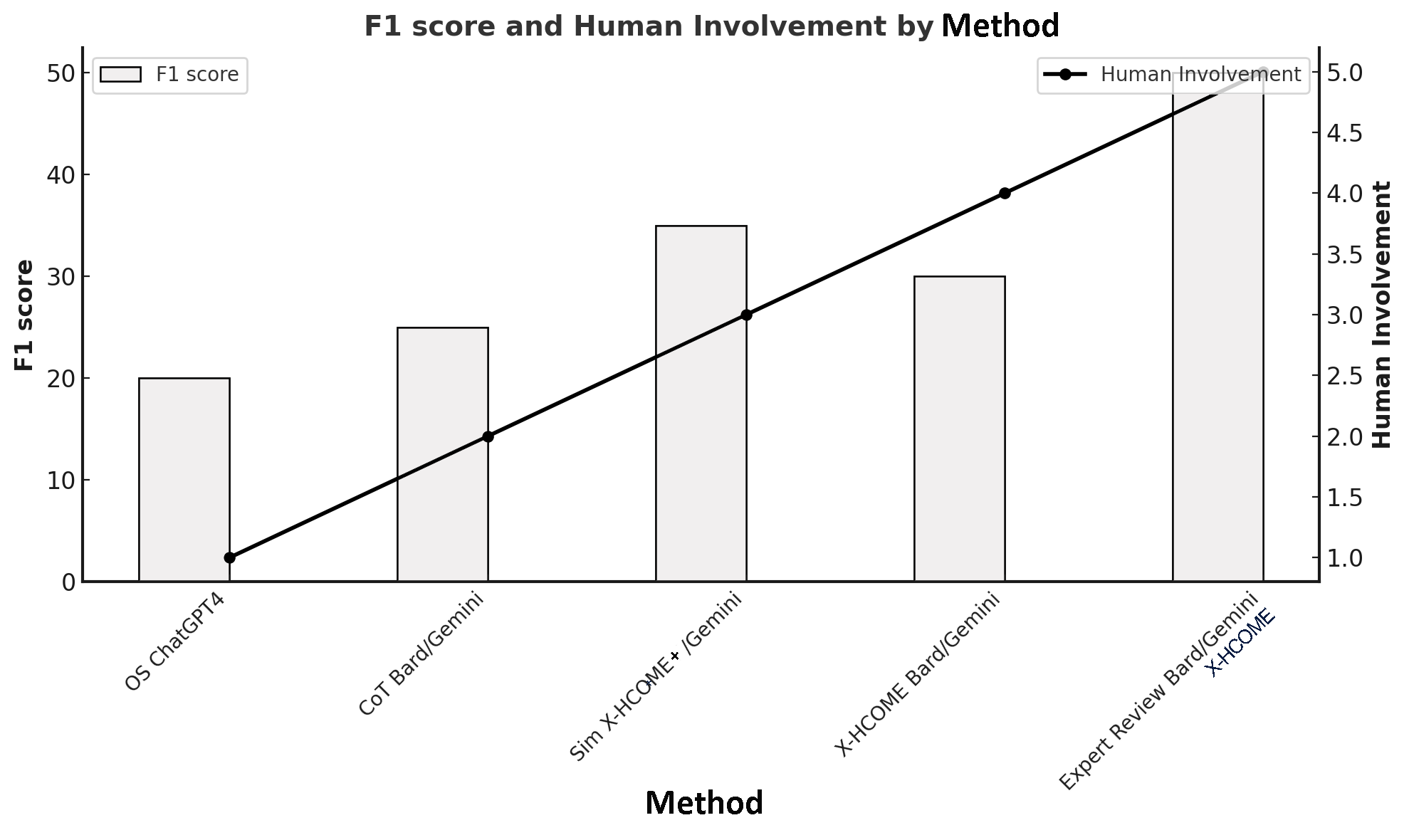}
    \caption{The graph compares the highest F1 scores from various LLM methods and the degree of human involvement in the PD domain. The x-axis represents the different methods: CoT Gemini, OS ChatGPT4, X-HCOME Bard/Gemini, Expert Review X-HCOME Bard/Gemini, and SimX-HCOME+ Gemini. The left y-axis shows the F1 score, while the right y-axis indicates the degree of human involvement, measured on a scale from 1 (minimum) to 5 (maximum).}
    \label{fig:human-inv}
\end{figure}

The results indicate that the Expert Review X-HCOME Bard/Gemini model achieved the highest F-1 score, exceeding 100\% and nearing 110\% with a human involvement level of 5. SimX-HCOME+ Gemini also showed a relatively high F-1 score, around 47.6\%, with a human involvement level of 3. X-HCOME Bard/Gemini had an F1 score slightly above 40\% with a human involvement level of 4. CoT Gemini and OS ChatGPT4 showed lower F-1 scores, around 20\%, with lower human involvement levels of 2 and 1 respectively (Figure~\ref{fig:human-inv}). These findings point to a positive relationship between the models' F-1 scores and the level of human engagement.

\section{Discussion}
The results presented in this paper partially confirm the first hypothesis that LLMs can autonomously develop an ontology for PD monitoring and alerting patients when provided with domain-specific input (aim, scope, requirements, competency questions, and data). While LLMs demonstrated the capability to construct an ontology, the comprehensiveness of this ontology did not fully align with the authors' expectations. LLMs efficiently acquired knowledge from big data repositories and generated ontologies using various prompt engineering techniques, but the resulting ontologies were not as comprehensive as anticipated.

On the other hand, the second hypothesis, which stated that combining human expertise with LLM capabilities improves the developed ontology's quality and comprehensiveness, was confirmed for PD monitoring and alerting of patients. The current paper demonstrates that the X-HCOME methodology provides a robust approach for developing comprehensive ontologies in the PD domain.

Regarding the third hypothesis, the results showed that expert revision can improve ontology generation, especially by reducing false positives. This kind of collaboration not only improves the structure and usefulness of the ontologies, but it may also uncover new information that expands the domain knowledge.

Also, concerning the paper's fourth hypothesis, the experiments conducted using the SimX-HCOME+ methodology further illustrate the importance of human-LLM collaboration. The inclusion of human experts in the iterative ontology generation and refinement process ensures that the ontologies produced are more comprehensive and detailed. However, regarding the transformation of NL to SWRL, the method did not fully manage to generate the SWRL rule, presenting a significant challenge for future experiments.

The conclusions of the paper emphasize that the three collaborative methods---X-HCOME, SimX-HCOME, and expert review of the X-HCOME---significantly enhance the comprehensiveness and time efficiency of ontology development. By integrating human expertise with the capabilities of LLMs, these methodologies address the limitations of LLMs in generating comprehensive ontologies independently.

However, collaborative methods like X-HCOME and SimX-HCOME+ may contain inherent biases in LLMs due to their training data and biases resulting from the opinions and experiences of specific domain experts. Another challenge with this paper is that it may have oversimplified the process of building an ontology by focusing too much on classes and object properties. Other crucial aspects, such as data properties and diverse axioms, are essential for crafting a comprehensive ontology. Finally, human evaluation is integral to the collaborative methodologies, and the comparison with the gold standard ontology and the use of metrics remain essential for validating and benchmarking the approach.

\section{Conclusions}
The promising results of the collaborative OEMs (X-HCOME, expert review of X-HCOME, and SimX-HCOME+) suggest their potential in further research efforts in LLM-enhanced OE, yet they also underscore the need for refinement before they can be considered mature OEMs. Future research could benefit from investigating the adaptability and effectiveness of these approaches in diverse healthcare contexts.

Regarding future work, it would be intriguing to explore the development of a specialized GPT model that is tailored specifically for ontology construction, utilizing the X-HCOME and SimX-HCOME+ methodologies.

%%%%%%%%%%% The bibliography starts:
% IMPORTANT for arXiv:
% - Upload bibliography.bib (must match the name below), OR
% - Provide a .bbl and replace the two lines below with: \input{nai_template.bbl}

\bibliographystyle{plain}
\bibliography{bibliography}

@misc{Ning2022,
  author = {Ning, Xiao and Celebi, Remzi},
  booktitle = {CEUR Workshop Proceedings},
  issn = {1613-0073},
  month = jan,
  pages = {46--54},
  title = {Knowledge Base Construction from Pre-trained Language Models by Prompt learning},
  url = {https://ceur-ws.org/Vol-3274/paper4.pdf},
  volume = {3274},
  year = {2022}
}

@article{Zafeiropoulos2023Wear4PDmove,
  author = {Zafeiropolos, Nikolaos and Bitilis, Pavlos and Kotis, Konstantinos},
  journal = {CEUR Workshop Proceedings},
  issn = {1613-0073},
  pages = {4},
  title = {Wear4pdmove: An Ontology for Knowledge-Based Personalized Health Monitoring of PD Patients},
  volume = {3632},
  year = {2023}
}

@article{Mountantonakis2023,
  author = {Mountantonakis, Michalis and Tzitzikas, Yannis},
  journal = {CEUR Workshop Proceedings},
  issn = {1613-0073},
  pages = {0--5},
  title = {Real-Time Validation of ChatGPT facts using RDF Knowledge Graphs},
  volume = {3632},
  year = {2023}
}

@misc{Oksanen2021,
  archivePrefix = {arXiv},
  arxivId = {2105.10966},
  author = {Oksanen, Joel and Cocarascu, Oana and Toni, Francesca},
  eprint = {2105.10966},
  title = {Automatic Product Ontology Extraction from Textual Reviews},
  url = {http://arxiv.org/abs/2105.10966},
  year = {2021}
}

@misc{Chowdhery2022,
  archivePrefix = {arXiv},
  arxivId = {2204.02311},
  author = {Chowdhery, Aakanksha and Narang, Sharan and Devlin, Jacob and Bosma, Maarten and Mishra, Gaurav and Roberts, Adam and Barham, Paul and Chung, Hyung Won and Sutton, Charles and Gehrmann, Sebastian and Schuh, Parker and Shi, Kensen and Tsvyashchenko, Sasha and Maynez, Joshua and Rao, Abhishek and Barnes, Parker and Tay, Yi and Shazeer, Noam and Prabhakaran, Vinodkumar and Reif, Emily and Du, Nan and Hutchinson, Ben and Pope, Reiner and Bradbury, James and Austin, Jacob and Isard, Michael and Gur-Ari, Guy and Yin, Pengcheng and Duke, Toju and Levskaya, Anselm and Ghemawat, Sanjay and Dev, Sunipa and Michalewski, Henryk and Garcia, Xavier and Misra, Vedant and Robinson, Kevin and Fedus, Liam and Zhou, Denny and Ippolito, Daphne and Luan, David and Lim, Hyeontaek and Zoph, Barret and Spiridonov, Alexander and Sepassi, Ryan and Dohan, David and Agrawal, Shivani and Omernick, Mark and Dai, Andrew M. and Pillai, Thanumalayan Sankaranarayana and Pellat, Marie and Lewkowycz, Aitor and Moreira, Erica and Child, Rewon and Polozov, Oleksandr and Lee, Katherine and Zhou, Zongwei and Wang, Xuezhi and Saeta, Brennan and Diaz, Mark and Firat, Orhan and Catasta, Michele and Wei, Jason and Meier-Hellstern, Kathy and Eck, Douglas and Dean, Jeff and Petrov, Slav and Fiedel, Noah},
  eprint = {2204.02311},
  month = apr,
  title = {PaLM: Scaling Language Modeling with Pathways},
  url = {http://arxiv.org/abs/2204.02311},
  year = {2022}
}

@article{Corra2023,
  author = {Corr{\`{a}}, Marta Francisca and Vila-Ch{\~{a}}, Nuno and Sardoeira, Ana and Hansen, Clint and Sousa, Ana Paula and Reis, In{\^{e}}s and Sambayeta, Firmina and Dam{\'{a}}sio, Joana and Calejo, Margarida and Schicketmueller, Andreas and Laranjinha, In{\^{e}}s and Salgado, Paula and Taipa, Ricardo and Magalh{\~{a}}es, Rui and Correia, Manuel and Maetzler, Walter and Maia, Lu{\'{i}}s F.},
  doi = {10.1093/brain/awac026},
  issn = {1460-2156},
  journal = {Brain},
  month = jan,
  number = {1},
  pages = {225--236},
  pmid = {35088837},
  publisher = {Oxford University Press},
  title = {Peripheral neuropathy in Parkinson's disease: prevalence and functional impact on gait and balance},
  volume = {146},
  year = {2023}
}

@article{Uschold1996,
  author = {Uschold, Mike and Gruninger, Michael},
  doi = {10.1017/S0269888900007797},
  issn = {1469-8005},
  journal = {The Knowledge Engineering Review},
  number = {2},
  pages = {93--136},
  publisher = {Cambridge University Press},
  title = {Ontologies: principles, methods and applications},
  url = {https://www.cambridge.org/core/journals/knowledge-engineering-review/article/abs/ontologies-principles-methods-and-applications/2443E0A8E5D81A144D8C611EF20043E6},
  volume = {11},
  year = {1996}
}

@misc{Sheth2023,
  archivePrefix = {arXiv},
  arxivId = {2305.00813},
  author = {Sheth, Amit and Roy, Kaushik and Gaur, Manas},
  eprint = {2305.00813},
  title = {Neurosymbolic AI -- Why, What, and How},
  url = {http://arxiv.org/abs/2305.00813},
  year = {2023}
}

@article{Bonuccelli2008,
  author = {Bonuccelli, Ubaldo and Ceravolo, Roberto},
  doi = {10.1517/14740338.7.2.111},
  issn = {1474-0338},
  journal = {Expert Opinion on Drug Safety},
  month = mar,
  number = {2},
  pages = {111--127},
  pmid = {18324875},
  title = {The safety of dopamine agonists in the treatment of Parkinson's disease},
  volume = {7},
  year = {2008}
}

@inproceedings{Bitilis2023,
  author = {Bitilis, Pavlos and Zafeiropoulos, Nikolaos and Koletis, Adam and Kotis, Konstantinos},
  booktitle = {14th International Conference on Information, Intelligence, Systems and Applications (IISA 2023)},
  doi = {10.1109/IISA59645.2023.10345958},
  isbn = {9798350318067},
  title = {Uncovering the Semantics of PD Patients' Movement Data Collected via off-the-shelf Wearables},
  year = {2023}
}

@article{Zafeiropoulos2023GNN,
  author = {Zafeiropoulos, Nikolaos and Bitilis, Pavlos and Tsekouras, George E. and Kotis, Konstantinos},
  doi = {10.3390/s23218936},
  issn = {1424-8220},
  journal = {Sensors},
  month = nov,
  number = {21},
  pages = {8936},
  pmid = {37960634},
  publisher = {MDPI},
  title = {Graph Neural Networks for Parkinson's Disease Monitoring and Alerting},
  url = {https://www.mdpi.com/1424-8220/23/21/8936},
  volume = {23},
  year = {2023}
}

@article{Younesi2015,
  author = {Younesi, Erfan and Malhotra, Ashutosh and G{\"{u}}ndel, Michaela and Scordis, Phil and Kodamullil, Alpha Tom and Page, Matt and M{\"{u}}ller, Bernd and Springstubbe, Stephan and W{\"{u}}llner, Ullrich and Scheller, Dieter and Hofmann-Apitius, Martin},
  doi = {10.1186/s12976-015-0017-y},
  issn = {1742-4682},
  journal = {Theoretical Biology and Medical Modelling},
  month = sep,
  number = {1},
  pmid = {26395080},
  publisher = {BMC},
  title = {PDON: Parkinson's disease ontology for representation and modeling of the Parkinson's disease knowledge domain},
  url = {https://link.springer.com/article/10.1186/s12976-015-0017-y},
  volume = {12},
  year = {2015}
}

@article{Kotis2006,
  author = {Kotis, Konstantinos and Vouros, George A.},
  doi = {10.1007/s10115-005-0227-4},
  issn = {0219-3116},
  journal = {Knowledge and Information Systems},
  number = {1},
  pages = {109--131},
  title = {Human-centered ontology engineering: The HCOME methodology},
  volume = {10},
  year = {2006}
}

@inproceedings{BouchourasGeorgiosBitilisPavlosKotisKonstantinos2024,
  author = {Bouchouras, Georgios and Bitilis, Pavlos and Kotis, Konstantinos and Vouros, George A.},
  booktitle = {GeNeSy2024 Workshop hosted by Extended Semantic Web Conference 2024},
  title = {LLMs for the Engineering of a Parkinson Disease Monitoring and Alerting Ontology},
  url = {https://sites.google.com/view/genesy2024/program},
  year = {2024}
}

@article{He2022,
  archivePrefix = {arXiv},
  arxivId = {2112.02682},
  author = {He, Yuan and Chen, Jiaoyan and Antonyrajah, Denvar and Horrocks, Ian},
  doi = {10.1609/aaai.v36i5.20510},
  eprint = {2112.02682},
  issn = {2159-5399},
  journal = {Proceedings of the AAAI Conference on Artificial Intelligence},
  pages = {5684--5691},
  title = {BERTMap: A BERT-Based Ontology Alignment System},
  volume = {36},
  year = {2022}
}

@article{Lippolis2023,
  author = {Lippolis, Anna Sofia and Klironomos, Antonis and Milon-Flores, Daniela F. and Zheng, Heng and Jouglar, Alexane and Norouzi, Ebrahim and Hogan, Aidan},
  journal = {CEUR Workshop Proceedings},
  issn = {1613-0073},
  title = {Enhancing Entity Alignment Between Wikidata and ArtGraph Using LLMs},
  url = {https://ceur-ws.org/Vol-3540/paper7.pdf},
  volume = {3540},
  year = {2023}
}

@misc{Funk2023,
  archivePrefix = {arXiv},
  arxivId = {2309.09898},
  author = {Funk, Maurice and Hosemann, Simon and Jung, Jean Christoph and Lutz, Carsten},
  eprint = {2309.09898},
  month = sep,
  title = {Towards Ontology Construction with Language Models},
  url = {http://arxiv.org/abs/2309.09898},
  year = {2023}
}

@article{Biester2023,
  author = {Biester, Fabian and Gaudio, Daniel Del and Abdelaal, Mohamed},
  journal = {CEUR Workshop Proceedings},
  issn = {1613-0073},
  title = {Enhancing Knowledge Base Construction from Pre-trained Language Models using Prompt Ensembles},
  url = {https://ceur-ws.org/Vol-3577/paper4.pdf},
  volume = {3577},
  year = {2023}
}

@misc{Pan2023,
  archivePrefix = {arXiv},
  arxivId = {2306.08302},
  author = {Pan, Shirui and Luo, Linhao and Wang, Yufei and Chen, Chen and Wang, Jiapu and Wu, Xindong},
  eprint = {2306.08302},
  title = {Unifying Large Language Models and Knowledge Graphs: A Roadmap},
  url = {http://arxiv.org/abs/2306.08302},
  year = {2023}
}

@misc{Joachimiak2009,
  author = {Joachimiak, Marcin P. and Caufield, J. Harry and Harris, Nomi L. and Kim, Hyeongsik and Mungall, Christopher J.},
  journal = {arXiv},
  title = {Gene Set Summarization using Large Language Models},
  year = {2009}
}

@misc{Caufield2023,
  archivePrefix = {arXiv},
  arxivId = {2304.02711},
  author = {Caufield, J. Harry and Hegde, Harshad and Emonet, Vincent and Harris, Nomi L. and Joachimiak, Marcin P. and Matentzoglu, Nicolas and Kim, HyeongSik and Moxon, Sierra A. T. and Reese, Justin T. and Haendel, Melissa A. and Robinson, Peter N. and Mungall, Christopher J.},
  eprint = {2304.02711},
  title = {Structured prompt interrogation and recursive extraction of semantics (SPIRES): A method for populating knowledge bases using zero-shot learning},
  url = {http://arxiv.org/abs/2304.02711},
  year = {2023}
}

@article{Mateiu2023,
  author = {Mateiu, Patricia and Groza, Adrian},
  doi = {10.1109/SYNASC61333.2023.00038},
  issn = {9798350394122},
  journal = {Proceedings - 2023 25th International Symposium on Symbolic and Numeric Algorithms for Scientific Computing (SYNASC)},
  month = jul,
  pages = {226--229},
  publisher = {IEEE},
  title = {Ontology engineering with Large Language Models},
  url = {https://arxiv.org/abs/2307.16699v1},
  year = {2023}
}

@article{Zeakis2023,
  author = {Zeakis, Alexandros and Papadakis, George and Skoutas, Dimitrios and Koubarakis, Manolis},
  doi = {10.14778/3598581.3598594},
  issn = {2150-8097},
  journal = {Proceedings of the VLDB Endowment},
  number = {9},
  pages = {2225--2238},
  title = {Pre-trained Embeddings for Entity Resolution: An Experimental Analysis},
  volume = {16},
  year = {2023}
}

@inproceedings{Jimenez-Ruiz2011,
  author = {Jim{\'{e}}nez-Ruiz, Ernesto and Cuenca Grau, Bernardo},
  booktitle = {Lecture Notes in Computer Science},
  doi = {10.1007/978-3-642-25073-6_18},
  isbn = {9783642250729},
  issn = {0302-9743},
  pages = {273--288},
  title = {LogMap: Logic-based and scalable ontology matching},
  volume = {7031},
  year = {2011}
}

\end{document}